\documentclass[conference]{IEEEtran}
\usepackage{graphicx}
\usepackage{subfigure}
\usepackage{multirow}
\usepackage{url}
\usepackage{amsmath}
\usepackage{amsopn}
\usepackage{url}

\graphicspath{{./Figures/}
}

\begin{document}
%
\title{Image-based Vehicle Analysis using Deep Neural Network: A Systematic Study}


\author{\IEEEauthorblockN{Yiren Zhou, Hossein Nejati, Thanh-Toan Do, Ngai-Man Cheung, Lynette Cheah}
\IEEEauthorblockA{
Singapore University of Technology and Design\\
yiren\_zhou@mymail.sutd.edu.sg,
\{hossein\_nejati, thanhtoan\_do, ngaiman\_cheung, lynette\}@sutd.edu.sg}
}


\maketitle

\begin{abstract}

We address the vehicle detection and classification problems using Deep Neural Networks (DNNs) approaches. Here we answer to questions that are specific to our application including how to utilize DNN for vehicle detection, what features are useful for vehicle classification, and how to extend a model trained on a limited size dataset, to the cases of extreme lighting condition.
Answering these questions we propose our approach that outperforms state-of-the-art methods, and achieves promising results on image with extreme lighting conditions.
\\[1\baselineskip]
\end{abstract}

\vspace{-6mm}
\begin{IEEEkeywords}
Vehicle Classification, Deep Neural Network
\end{IEEEkeywords}

%
\IEEEpeerreviewmaketitle

\vspace{-1.5 mm}
\section{Introduction}
\label{sec:introduction}

Vehicle detection and classification are important parts of Intelligent Transportation Systems.
They aid traffic monitoring, counting, and surveillance, which are necessary for tracking the performance of traffic operations. Existing methods use various types of information for vehicle detection and classification, including acoustic signature~\cite{wang2014vehicle},
radar signal~\cite{kim2013vehicle},
frequency signal~\cite{mckay2012passive},
and image/video representation~\cite{mishra2013multiple}.
The evolution of image processing techniques, together with wide deployment of road cameras, facilitate image-based vehicle detection and classification.

Various approaches to image-based vehicle detection and classification have been proposed recently.
Sivaraman and Trivedi~\cite{sivaraman2012real}
use active learning to learn from front part and rear part vehicle images, and achieves 88.5\% and 90.2\% precision respectively.
Chen et al.~\cite{chen2012vehicle} use a combination of Measurement Based Features~(MBF) and intensity pyramid-based HOG~(IPHOG) for vehicle classification on front view road images.
A rear view vehicle classification approach is proposed by Kafai and Bhanu~\cite{kafai2012dynamic}. They define a feature set including tail light and plate position information, then pass it into hybrid dynamic Bayesian network for classification. 

Fewer efforts have been devoted in 
rear view vehicle classification~\cite{kafai2012dynamic}.  Rear view vehicle classification is an important problem as many road cameras capture rear view images. Rear views are also less discriminative and therefore more challenging.
Furthermore, it is more  challenging for images captured from a distance along multi-lane highways, with possiblity of partial occlusions and motion blur that complicate detection and classification.

We here focus on
DNN-based vehicle detection and classification based on rear view images, captured by a static road camera from a distance along a multi-lane highway (Fig.~\ref{fig:road_image}).
DNN has been applied to many image/video applications~\cite{girshick2015fast,krizhevsky2012imagenet,sibo:2016,toan:2016,victor:2016}.
Whilet these methods achieve state-of-the-art on various datasets~\cite{razavian2014cnn}, direct application of them requires a large dataset, that is laborious and expensive to construct. Training of DNN on a small dataset on the other hand, would result in overfitting. Given the difficulty of the original problem (i.e. large in-class variances and ambiguity), reliable modeling based on a small dataset proves even more challanging.

In this work, we propose a combination of approaches to use DNN architectures for this specific problem, building around using the higher layers of a DNN trained on a specific large labeled dataset~\cite{yosinski2014transferable}.
There are two approaches to making use of the higher layers of DNN architecture: one is to fine-tune the higher layers of DNN model on our dataset, and the other is to extract the higher layers of DNN architecture as high-level features, and use them for detection and classification.
Proposed schemes for our approach are shown in Table~\ref{tab:scheme}.

Comparing with the state-of-the-art classification methods shows that the vehicle classification methods achieve the highest accuracies. In addition, when coupled with illumination and color transformation and late fusion, the same model retain robustness in  classification of poorly lit images $without$ fine-tuning or re-training the classification model. Without color transformation, these dark images significantly affect classification results. Our contribution is therefore an approach to train DNN models for vehicle detection and classification on small datasets, and extend their application to cases beyond the content of the original small dataset.

\begin{table*}[t]
\centering
\begin{tabular}{|c|c|c|}
\hline
   & Scheme & Dataset~\cite{dataset} \\
   \hline
\multirow{2}{*}{Vehicle detection} & Fine-tune YOLO model & \multirow{2}{*}{\begin{tabular}[c]{@{}c@{}}A: 438 road images,\\ 263 for training, 175 for testing\end{tabular}}   \\ \cline{2-2}
& Conventional state-of-the-art & \\
\hline
\multirow{3}{*}{Vehicle classification} & Fine-tune Alexnet model & \multirow{3}{*}{\begin{tabular}[c]{@{}c@{}}B: 2427 vehicle images,\\ 1440(845 passenger, 595 other) for training,\\ 987(597 passenger, 390 other) for testing\end{tabular}} \\ \cline{2-2} & Alexnet feature extraction  & \\
\cline{2-2}
 & Conventional state-of-the-art & \\
 \hline
\multirow{3}{*}{\begin{tabular}[c]{@{}c@{}}Vehicle classification\\ on dark images\end{tabular}} & Classification on dark image          & \multirow{3}{*}{\begin{tabular}[c]{@{}c@{}}C: 257 dark vehicle images,\\ 223 passenger, 34 other\end{tabular}} \\
\cline{2-2}
& Classification on transformed image   & \\ \cline{2-2}
 & Late-fusion & \\
 \hline
\end{tabular}
\vspace{0.5mm}
\caption{Proposed schemes for our approach.}
\label{tab:scheme}
\vspace{-9 mm}
\end{table*}

%
\vspace{-1.5 mm}
\section{Methodology}
\label{sec:methodology}



\subsection{Vehicle detection using YOLO}
\label{ssec:detection}
\vspace{-1 mm}

\begin{figure}[htbp]
\vspace{-4mm}
\centering
\subfigure[]{
       \includegraphics[width=0.45\columnwidth]{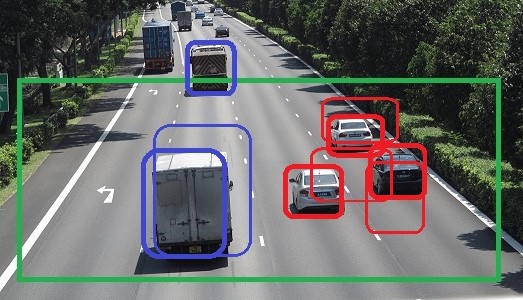}
       \label{fig:road_image_3}
}
\subfigure[]{
       \includegraphics[width=0.45\columnwidth]{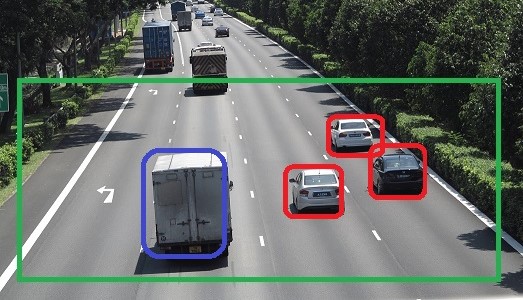}
       \label{fig:road_image_4}
}
\vspace{-3 mm}
\caption{Examples for vehicle detection approach on a road image. The green rectangle is the selected road region for detection. Red and blue rectangles in (a) are the initial detection results by YOLO model. After remove invalid detection results, the final detection results are shown in (b).}
\label{fig:road_image}
\vspace{-4 mm}
\end{figure}

Our dataset images are taken from a static camera along an express way and contain rear views of vehicles on multiple lanes~(Fig.~\ref{fig:road_image}). 
We manually label the location bounding boxes for vehicles inside each the road region closest to the camera. A detected vehicle object is valid only if the center of the object is inside the selected road region.

DNN architecture has been widely applied for object detection tasks. Fast R-CNN~\cite{girshick2015fast} achieved state-of-the-art result on several datasets. However, a more recent DNN method called YOLO~\cite{redmon2015you} achieved comparable results while being significantly faster. To do the vehicle detection more efficiently, we choose YOLO for our approach.

\begin{figure}[htbp]
\centering
\includegraphics[width=\columnwidth]{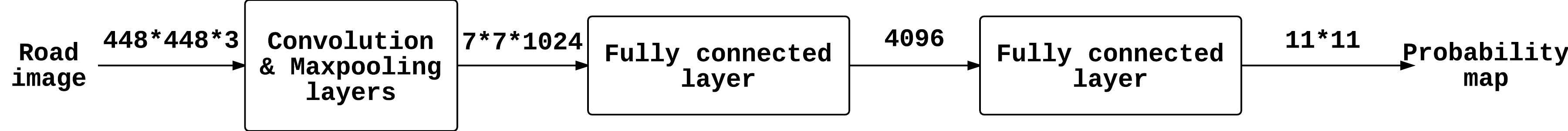}
\caption[]{Simplified YOLO network structure. A detailed structure can be found in~\cite{redmon2015you}.}
\label{fig:yolo}
\end{figure}

Fig.~\ref{fig:yolo} shows the simplified structure for YOLO network.
The original YOLO network is trained on PASCAL dataset~\cite{pascal-voc-2012} with 20 classes of objects with a probability grid with size \textit{7*7*20}. We can increase the size of probability grid to improve detection accuracy. However, it will also increase model complexity, and require higher number of training samples. Here we increase the probability grid to \textit{11*11}, and to number of classes to 1, resulting in a probability grid with size \textit{11*11}. Then we fine-tune the last layer of the model with our own road images.

Fig.~\ref{fig:road_image_3} shows that the fine-tuned YOLO model has generates some invalid detection results. We therefore use a post-processing approach to remove these outliers. Outliers are removed based on the following criterion:
\[
\text{$A$ is}
\begin{cases}
invalid, & \text{if } \exists B \in \text{image, } (\frac{Int(A, B)}{Area(A)}>t \| \frac{Int(A, B)}{Area(B)}>t)\\ 
&\& Conf(A) < Conf(B) \\
invalid, & \text{else if } Center(A) \notin Region(valid)\\
valid, & \text{otherwise}
\end{cases}
\]
where $A,B$ are two different detected bounding boxes. $Int(A,B)$ is the intersection area of $A,B$, $Area(A)$ is the area of $A$, $t$ is a threshold value, and $Conf(A)$ is the confidence value of $A$ given by YOLO model.
$Center(A)$ is the center pixel of $A$, and $Region(valid)$ represents the green rectangle shown in Fig.~\ref{fig:road_image_3}. The final detection results after post-processing are shown in Fig.~\ref{fig:road_image_4}.

The YOLO model can also perform vehicle classification when we set the number of class to 2, representing passenger and other vehicles. However, the classification accuracy for YOLO is not high enough. We continue to introduce more classification approaches in Section~\ref{ssec:classification}.
\vspace{-1.5 mm}
\subsection{Vehicle classification approaches}
\label{ssec:classification}
\vspace{-1 mm}

\begin{figure}[htbp]
\vspace{-4.5 mm}
\centering
\subfigure[]{
       \includegraphics[width=0.2\columnwidth]{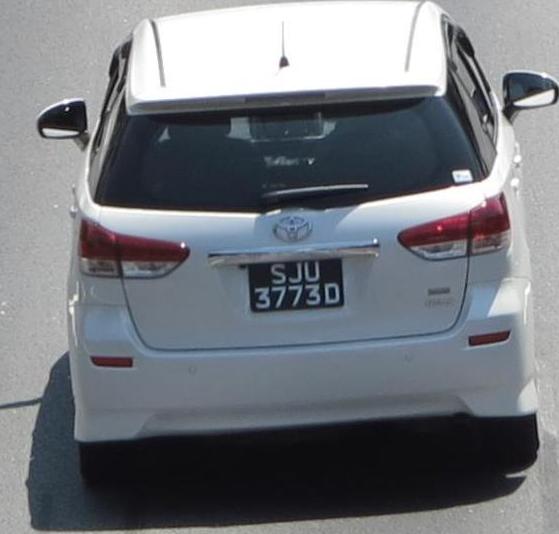}
       \label{fig:passenger_1}
}
\subfigure[]{
       \includegraphics[width=0.2\columnwidth]{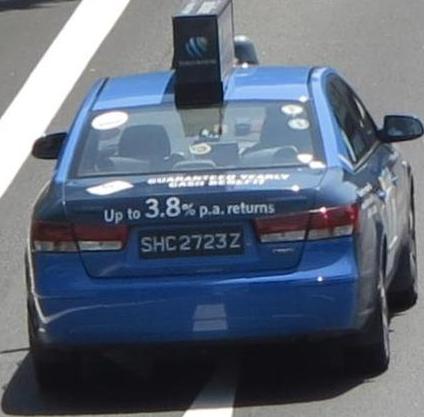} 
       \label{fig:passenger_3}
}
\subfigure[]{
       \includegraphics[width=0.2\columnwidth]{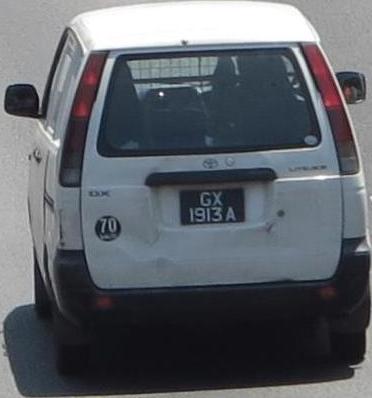}
       \label{fig:other_1}
}
\subfigure[]{
       \includegraphics[width=0.2\columnwidth]{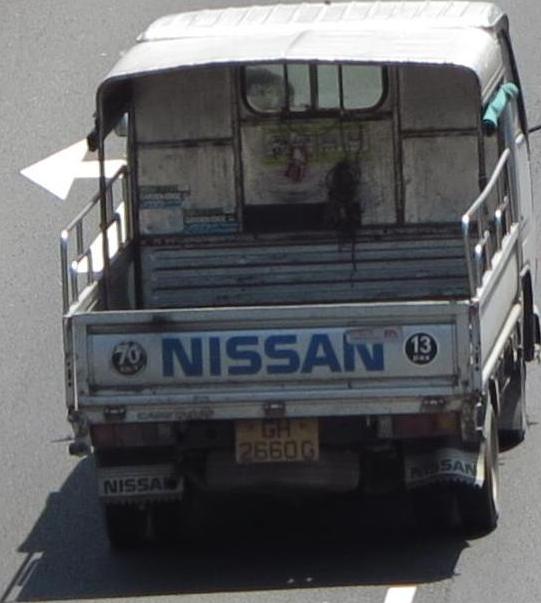}
          \label{fig:other_2}
}
\\
\vspace{-3 mm}
\caption[]{Vehicle image examples for both classes. (a) passenger. (b) passenger.
(c) other. (d) other.
}
\label{fig:sample_vehicle}
\vspace{-4 mm}
\end{figure}

For vehicle classification, we use the dataset B described in Table~\ref{tab:scheme}.
Vehicle images will be classified into two classes: {\em passenger} and {\em other}. Passenger vehicle class includes sedan, SUV, and MPV, other vehicle class includes van, truck, and other types of vehicle. Both classes have large in-class variance. Also the difference between passenger vehicles and other vehicles is not distinctive. These make it difficult to distinguish between these two classes. Fig.~\ref{fig:sample_vehicle} shows examples for both vehicle classes. As we can see from the sample images, Fig.~\ref{fig:passenger_1} is MPV, and Fig.~\ref{fig:passenger_3} is taxi. They are both passenger vehicles but different in shape, color, and size. Fig.~\ref{fig:passenger_1} is MPV, and Fig.~\ref{fig:other_1} is van. They are in different classes, but similar in shape, color, and size. The classification between passenger vehicles and other vehicles has semantic meanings included, that can only be represented using both low-level and high-level features.

Here we apply two approaches for utilizing DNN architecture: feature extraction, and fine-tuning. For both approaches, we adopt Alexnet~\cite{krizhevsky2012imagenet} model as DNN architecture.

\begin{figure}[htbp]
\vspace{-3 mm}
\centering
\includegraphics[width=0.8\columnwidth]{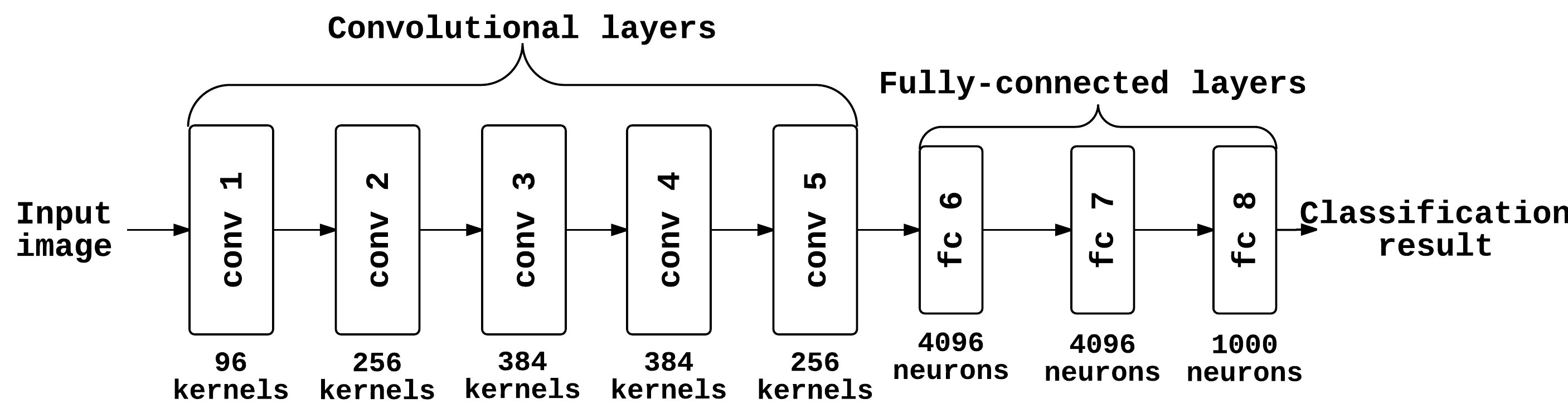}
\vspace{-4 mm}
\caption[]{Structure of Alexnet.}
\label{fig:alexnet}
\vspace{-3 mm}
\end{figure}

For each vehicle image detected from Section~\ref{ssec:detection}, we resize it to 256~$\times$ 256, make it valid Alexnet input. Then the resized image is passed into Alexnet. Fig.~\ref{fig:alexnet} shows the structure of Alexnet. Alexnet has 5 convolutional layers~(named as conv1 to conv5) and 3 fully-connected layers~(named as fc6, fc7, fc8). Each convolutional layer contains multiple kernels, and each kernel represents a 3-D filter connected to the outputs of the previous layer. For fully-connected layers, each layer contains multiple neurons. Each neuron contains a positive value, and it is connected to all the neurons in previous layer.

\subsubsection{Feature extraction using Alexnet}


Here we extract the third last and second last fully connected layer~(i.e. layer fc6 and fc7) in Alexnet as the generic image representation (to be justified later). Each image representation is a 4096-dimension vector, obtained from the 4096 neurons in layer fc6~(or fc7). Here we consider the extracted layer as a feature vector $f=[f_1,f_2,...,f_{4096}]$.
After we obtain the 
deep feature vector,
SVM with linear kernel is used for classification.

\begin{figure}[htbp]
\vspace{-3 mm}
\centering
\subfigure[]{
       \includegraphics[width=0.2\columnwidth]{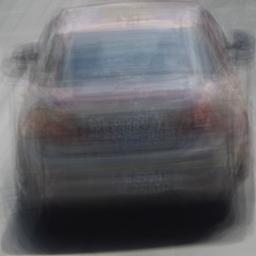}
       \label{fig:mean_1}
}
\subfigure[]{
       \includegraphics[width=0.2\columnwidth]{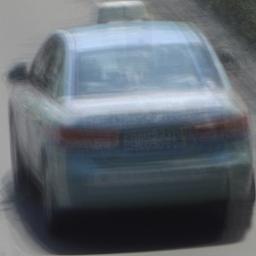}
       \label{fig:mean_4}
}
\subfigure[]{
       \includegraphics[width=0.2\columnwidth]{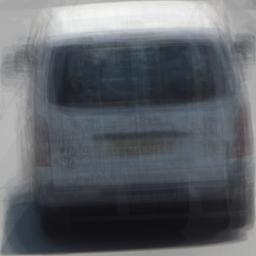}
       \label{fig:mean_6}
}
\subfigure[]{
       \includegraphics[width=0.2\columnwidth]{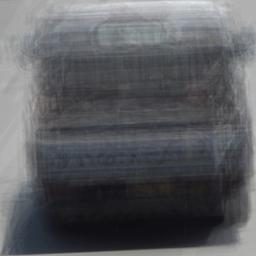}
       \label{fig:mean_8}
}\\
\vspace{-3 mm}
\caption[]{Average image of the vehicles with high values on a specific feature.}
\label{fig:mean-image}
\vspace{-3 mm}
\end{figure}

Different layers in a Deep Neural Network~(DNN) are often considered to have different level of features. The first few layers contain general features that resemble Gabor filters or blob features. The higher layers contain specific features, each representing a particular class in dataset~\cite{yosinski2014transferable}. Thus features in higher layers are considered to have higher level vision information compared to general features in base layers. 
To understand this in our particular problem, Fig.~\ref{fig:mean-image} shows several average images we obtained from vehicle images. Given a specific feature $f_i$ we extracted from Alexnet, we sort all the vehicle images based on value of $f_i$. The images that have highest values on this feature are chosen. Then we calculate the average image of these images. 
The 4 images in Fig.~\ref{fig:mean-image} represents average images for 4 different features~(i.e. $f_{i_1},...,f_{i_4}$, here $i_1,..,i_4 \in \{1,...,4096\}$). We can recognize specific types of vehicles from these average images. Fig.~\ref{fig:mean_1} represents a specific type of normal sedan. Fig.~\ref{fig:mean_4} is taxi. Fig.~\ref{fig:mean_6} is van. And Fig.~\ref{fig:mean_8} represents truck. Human can easily associate these average images to certain types of vehicles, meaning that the features related to these images contain high-level visualization information related to semantic meanings of each class.

\subsubsection{Fine-tuning Alexnet on our dataset}

Another approach to make use of the high-level information in DNN is to fine-tune the DNN model on our dataset. Alexnet~\cite{krizhevsky2012imagenet}
is trained
with 1000 classes.
To match our dataset with
2 classes, we change the size of fc8 layer of Alexnet from 1000 to 2. Then we use Alexnet model trained on ILSVRC 2012 dataset to fine-tune on our dataset. In order to prevent overfitting, the parameters from layer conv1 to layer fc6 is fixed. After the fine-tuning, the model is tested on testing set with 987 images.

\vspace{-1.5mm}
\subsection{Vehicle classification on dark images}

\begin{figure}[htbp]
\vspace{-3mm}
\centering
\subfigure[]{
       \includegraphics[width=0.21\columnwidth]{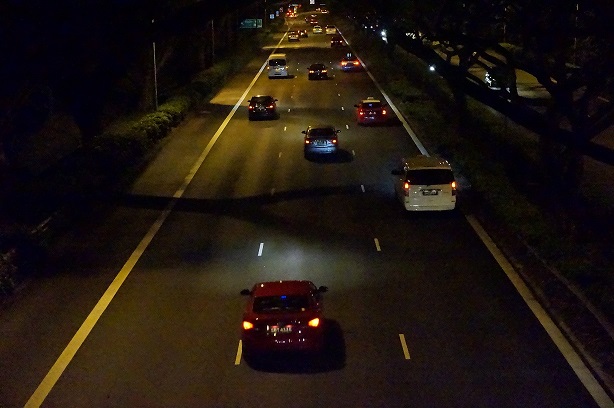}
       \label{fig:night}
}
\subfigure[]{
       \includegraphics[width=0.21\columnwidth]{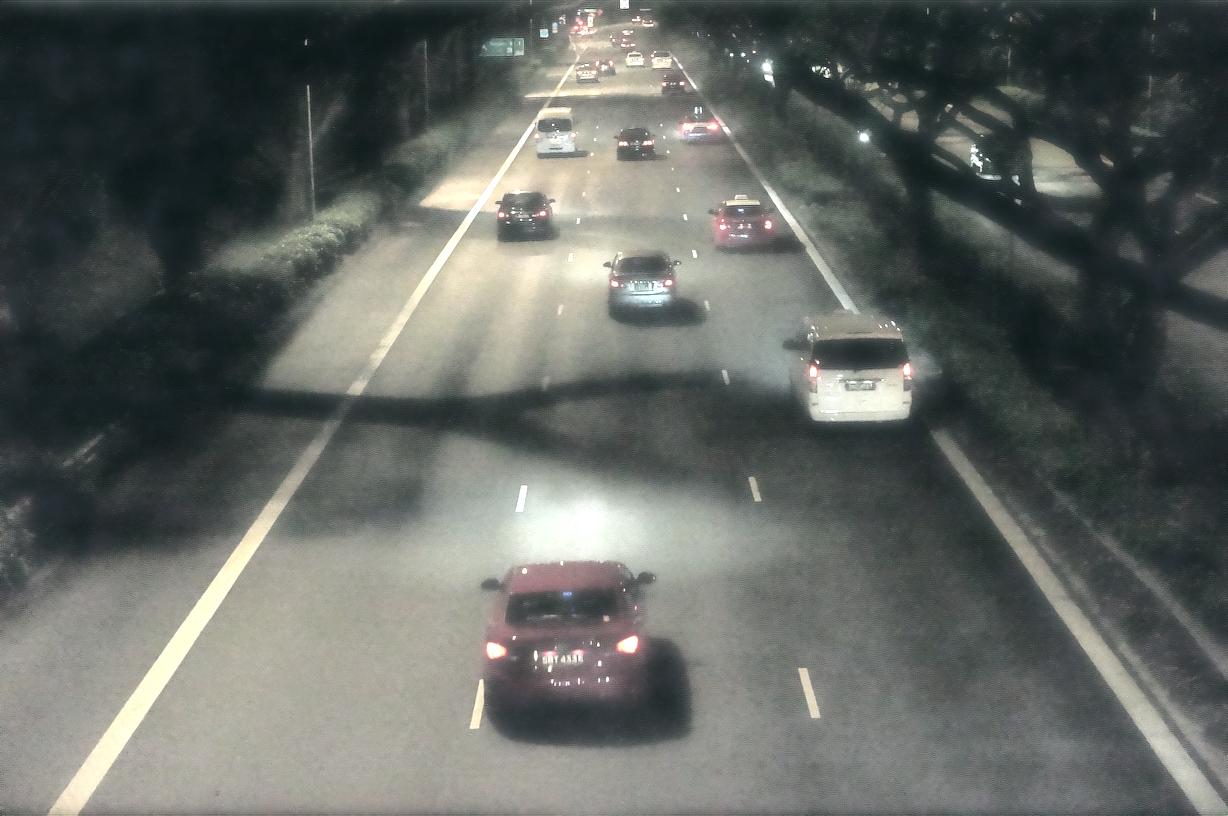}
       \label{fig:night-tran}
}
\subfigure[]{
       \includegraphics[width=0.18\columnwidth]{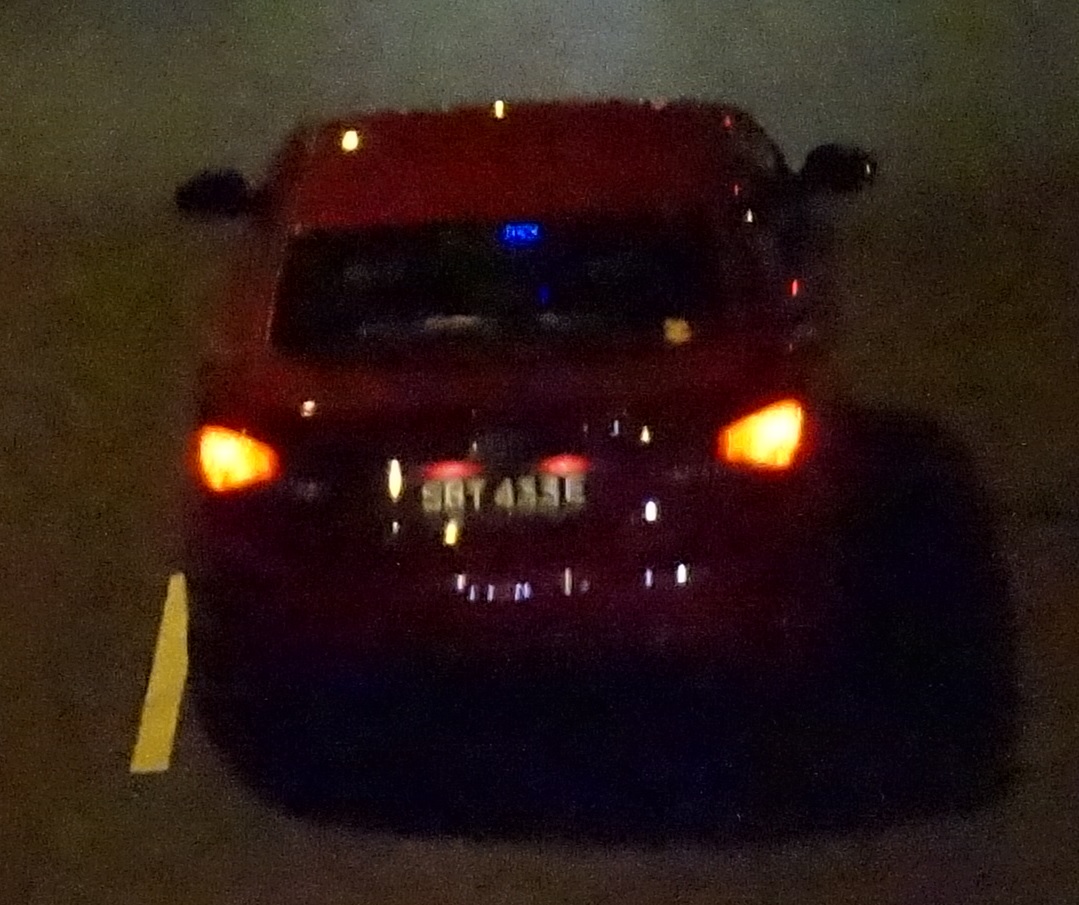}
       \label{fig:dark-vehicle}
}
\subfigure[]{
       \includegraphics[width=0.18\columnwidth]{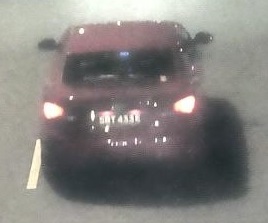}
       \label{fig:dark-vehicle-tran}
}
\vspace{-3mm}
\caption[]{Dark image examples.}
\label{fig:dark-image}
\vspace{-3mm}
\end{figure}

There is also a need for vehicle classification on dark images. Fig.~\ref{fig:night} shows an image taken during the night, where classification is more challenging due to poor lighting. One approach to improve accuracy for dark vehicle image classification is to train the model on dark images, however, it is not feasible when we have a limited number of dark image samples. Here we propose a method to use model trained on normal images to classify dark images.

Dark image often comes with low contrast, color displacement, and high noise that would significantly affect image quality. By applying color transformation method we can increase image contrast and fix color displacement. In addition, the classification model is trained on \textit{normal} images, so we want to have a transformed image close to \textit{normal} images. Here we make use of a high-level scene transformation method~\cite{Laffont14} to transform \textit{night} image back to \textit{normal} image. This transformation model are trained from scene image dataset with different lighting, weather, and seasons. It can conduct high-level scene transformation, including the transformation between different lighting conditions.

Fig.~\ref{fig:night-tran} is the transformed image by using~\cite{Laffont14}. If we see details in passenger vehicle Fig.~\ref{fig:dark-vehicle}, the contrast level is low, and there also exist color displacement and noise. By using scene transformation, the contrast level of Fig.~\ref{fig:dark-vehicle-tran} increases, color displacement and noise still remains. However, from the transformed image Fig.~\ref{fig:dark-vehicle-tran} we can observe that the color displacement is different from Fig.~\ref{fig:dark-vehicle}. From these results we propose a late fusion method to utilize both the dark image and the transformed image for classification.

The late fusion method is listed as follows:

\vspace{-4mm}
\begin{equation}
\begin{split}
Label_{fused}=\arg\underset{i}{\max}\,&Conf(i,j), i\in \{passenger, other\} \\
& j\in \{original, transformed\}
\label{eq:fusion}%
\end{split}
\end{equation}
\vspace{-3mm}

where $Conf(i,j)$ is the confidence score of class $i$ from image $j$, generated from SVM model. For Alexnet fine-tuning model, the confidence score is generated by the softmax layer. The idea is to select the classification result with highest confidence on both night image and transformed image.

We test the scene transformation and late fusion methods on a night image dataset C in Table~\ref{tab:scheme}.
The test results are reported in Section~\ref{sssec:experiment_our_dataset}.
\vspace{-1.5 mm}
\section{Experimental results}
\label{sec:experiment}

In this section the experimental results of the proposed vehicle detection and vehicle classification method are presented. The size of road images in our dataset is 4184~$\times$ 3108. For the vehicle detection process, the road images are resized to 448~$\times$ 333 using Bicubic interpolation. After vehicle detection, we map the vehicle regions back to original road images, and crop vehicle image in original resolution. Typical resolution of vehicle images is around 500~$\times$ 500. For vehicle classification, all vehicle images are resized to 256~$\times$ 256 to pass into Alexnet, for other embedding methods, we use vehicle images with original resolution. The vehicle detection method is implemented in darknet~\cite{darknet}. For vehicle classification method, the feature extraction and fine-tuning of Alexnet is under Caffe framework~\cite{jia2014caffe}, other feature embedding methods and SVM are implemented in MATLAB.

\vspace{-2.5 mm}
\subsection{Vehicle detection experiment}
\label{ssec:experiment_detection}
\vspace{-1 mm}


\begin{table}[htbp]
\vspace{-3mm}
\centering
\begin{tabular}{|c|c|cc|cc|}
\hline
\multicolumn{2}{|c|}{\multirow{2}{*}{Detection result}} & \multicolumn{2}{c|}{YOLO fine-tune} & \multicolumn{2}{c|}{DPM~\cite{lsvm-pami}} \\ \cline{3-6} 
\multicolumn{2}{|c|}{}                                  & positive         & negative         & positive    & negative   \\ \hline
\multirow{2}{*}{Ground truth}         & positive        & 921              & 185              & 932         & 149        \\
                                      & negative        & 66               & -                & 55          & -          \\ \hline
\end{tabular}
\vspace{0.5mm}
\caption{Comparison of detection result.}
\label{tab:detection}
\vspace{-6mm}
\end{table}

We train and test the YOLO detection model on dataset A described in Table~\ref{tab:scheme}. Here we compare the result with another state-of-the-art detection method DPM~\cite{voc-release5,lsvm-pami}. 
Among 987 testing images, 921 are successfully detected, 185 detected images are invalid images. Vehicle detection precision is 93.3\%, and recall is 83.3\%, compared with 94.4\% and 86.2\% by DPM method.

\subsection{Vehicle classification experiment}
\label{ssec:experiment_classification}

\subsubsection{Experiment on public dataset}

\begin{table}[htbp]
\vspace{-3 mm}
  \centering
    \begin{tabular}{|c|ccc|}
    \cline{1-4}
    \multirow{2}[0]{*}{Accuracy (\%)} & Cars vs & Sedans vs & Sedans vs vans \\
          & vans & taxis & vs taxis \\
    \cline{1-4}
    PCA+DFVS~\cite{ambardekar2014vehicle} & 98.50  & \textbf{97.57} & 95.85 \\
    PCA+DIVS~\cite{ambardekar2014vehicle} & 99.25 & 89.69 & 94.15 \\
    PCA+DFVS+DIVS~\cite{ambardekar2014vehicle} &   -   &   -   & 96.42 \\
    Constellation model~\cite{ma2005edge} & 98.50  & 95.86 &  \\
    \cline{1-4}
    Alexnet-fc6-SVM & \textbf{99.50}  & 97.27 & \textbf{97.36} \\
    Alexnet-fc7-SVM & 99.25  & 96.67  & 94.75 \\
    \cline{1-4}
    \end{tabular}%
    \vspace{0.5 mm}
  \caption{Accuracy comparison on public dataset. Reported results from~\cite{ambardekar2014vehicle} are used.}
  \label{tab:accuracy_table_public_dataset}%
  \vspace{-6 mm}
\end{table}%

To compare our approach with 
other
classification methods, we perform our approach on a public dataset provided in~\cite{ma2005edge}.
We use same experiment setting in~\cite{ambardekar2014vehicle} to perform fair comparison. There are three types of vehicles in this dataset: sedans, vans, and taxis. Following~\cite{ambardekar2014vehicle}, three experiments are performed: \textit{cars} vs \textit{vans}, \textit{sedans} vs \textit{taxis}, and \textit{sedans} vs \textit{vans} vs \textit{taxis}. Note that sedans and taxis are all regarded
as cars.

We apply feature extraction on Alexnet model for classification. From each vehicle images, we extract two feature vectors~(from layer fc6 and fc7) with 4096 dimensions using Alexnet. Then, linear-SVM with is applied for classification.


Table~\ref{tab:accuracy_table_public_dataset} shows accuracy comparison among Alexnet-based methods and other state-of-the-art methods.
Alexnet-fc6 feature achieves best accuracy on \textit{cars} vs \textit{vans}, and \textit{sedans} vs \textit{vans} vs \textit{taxis} classification, and second-best accuracy on \textit{sedans} vs \textit{taxis} classification. These results show the effectiveness of Alexnet features on vehicle classification problem.
\subsubsection{Experiment on our dataset}
\label{sssec:experiment_our_dataset}

The vehicle classification approaches are trained and tested on dataset B\footnote{The dataset can be downloaded via link~\cite{dataset}.} in Table~\ref{tab:scheme}.
All results are calculated using class-balanced accuracy
as shown below:

\vspace{-4mm}
\begin{equation}
Acc_{bal}=\frac{\frac{Correct(pass)}{Size(pass)}+\frac{Correct(other)}{Size(other)}}{2}
\label{eq:balance-acc}
\end{equation}
\vspace{-4mm}

where $Correct(pass)$ is the number of correct prediction in passenger class, and $Size(pass)$ is the total number of images in passenger class.



Here we compare the performance of the DNN with state-of-the-art image description methods: Fisher vector~\cite{sanchez2013image}, FAemb~\cite{do2015faemb,Do_2016}, and Temb~\cite{jegou2014triangulation} with SIFT descriptor\footnote{Unable to run the code from~\cite{ambardekar2014vehicle}, we did not include their methods.}.

From each vehicle image, we extract a feature vector~(fc6 or fc7) with 4096 dimensions using Alexnet. Another alternative method is to concatenate fc6 and fc7 to get 8192 dimensions feature vector. For Alexnet fine-tuning, we directly use the fine-tuned model to classify vehicle images.

For comparison with state-of-the-art methods, from each vehicle image, we first compute SIFT descriptors (each having 128 dimensions) of the image.
Then different embedding features are generated based on SIFT.
Generated fisher vectors have 4k or 8k dimensions. Temb and Faemb have around 8k dimensions.


For both Alexnet extracted features and other methods, we use linear-SVM to train the classifier.
The SVM is trained on dataset B in Table~\ref{tab:scheme}.
The trained model is also tested on a dark image dataset C.



\begin{table}[htbp]
  \vspace{-3 mm}
  \centering
    \begin{tabular}{|c|c|c|ccc|}
\hline
\multirow{2}[0]{*}{Accuracy (\%)} &       & Normal & Dark   & Transformed & Late    \\
                  & Dims  & images & images & images      & fusion  \\
                  \hline
Fisher-vec-4k~\cite{sanchez2013image} & 4096  & 93.55  & 57.3   & 60.06       & 60.85   \\
Fisher-vec-8k~\cite{sanchez2013image} & 8192  & 93.3   & 58.98  & 62.37       & 61.87   \\
FAemb~\cite{do2015faemb,Do_2016}  & 8280   & 89.76  & 69.7   & 60.13       & 68.67   \\
Temb~\cite{jegou2014triangulation} & 8192  & 87.81  & 65.35  & 56.81       & 64.55   \\
\hline
Alexnet-fc6       & 4096  & 96.95  & \textbf{74.68}  & \textbf{79.32}       & \textbf{85.41}   \\
Alexnet-fc7       & 4096  & 96.44  & 57.93  & 64.91       & 67.51   \\
Alexnet-fc6\&7    & 8192  & \textbf{97.35}  & 73.65  & 77.97       & 84.16   \\
Alexnet-fine-tune & -     & 97.15  & 52.76  & 52.08       & 52.52   \\
\hline
\end{tabular}%
    \vspace{0.5 mm}
  \caption{Accuracy comparison on our dataset.}
  \label{tab:accuracy_table_our_dataset}%
  \vspace{-6 mm}
\end{table}%

Table~\ref{tab:accuracy_table_our_dataset} shows the accuracy comparison for all methods. The best result is achieved by concatenating fc6 and fc7 layers of Alexnet. Fine-tuning also achieves good result for classification on normal images.

For dark image results, all methods have suffered from severe accuracy degradation. fc6 feature has achieved best result on dark images. We can see the fc7 feature has much lower result compared to fc6, indicating that fc6 is more robust to low contrast and color displacement. It is also interesting to see that fine-tune Alexnet model has poor result on dark images. The fc7 model and fine-tuned Alexnet model is fitted into normal image classification, and cannot generalize to dark image classification.

All Alexnet features have improvement on transformed images, indicating the effectiveness of scene transformation on dark images for classification. Other state-of-the-art feature embedding methods are not benefited from the transformation, because these features are generated from SIFT feature, and SIFT feature does not have strong relationship with color transformation of the image.

The late fusion results are shown in the last column. We can see that the Alexnet features have improvements after we conduct late fusion on dark and transformed images. The best result is achieved by fc6 feature for about 85\%.
\vspace{-1.5 mm}
\section{Conclusion}
\label{sec:conclusion}

We have investigated DNN approaches for both vehicle detection and classification using a limited size dataset.
For detection, we fine-tune a DNN detection model for vehicle detection and achieved good result. For classification, we evaluate both fine-tuning and feature-extraction method, the result outperformed state-of-the-art.

We further proposed methods to use scene transformation and late fusion techniques for classification on poor lighting conditions, and achieved promising results without changing the classification model. Our approach is therefore have the potential to be used for training on limited size datasets and be extended to different cases such as various lighting conditions.


\bibliographystyle{IEEEBib}
\bibliography{refs}

\end{document}